\documentclass[journal]{IEEEtran}
\usepackage{times}
\usepackage{epsfig}
\usepackage[caption=false,font=footnotesize]{subfig}
\usepackage{adjustbox}
\usepackage[linesnumbered,ruled,vlined]{algorithm2e}
\usepackage{caption}
\usepackage{multirow}
\usepackage{bm}
\usepackage{esvect}
\usepackage[normalem]{ulem}
\useunder{\uline}{\ul}{}
\usepackage[dvipsnames]{xcolor}
\usepackage{setspace}
\usepackage{amsmath}
\usepackage{breqn}
\usepackage{booktabs}
\usepackage[pagebackref=true,breaklinks=true,letterpaper=true,colorlinks,bookmarks=false]{hyperref}
\usepackage{graphicx}
\begin{document}

%%%%%%%%% TITLE
\title{Deriving Explanation of Deep Visual Saliency Models}

\author{\IEEEauthorblockN{Sai Phani Kumar Malladi\IEEEauthorrefmark{1},
Jayanta Mukhopadhyay\IEEEauthorrefmark{1},
Chaker Larabi\IEEEauthorrefmark{2},
Santanu Chaudhury\IEEEauthorrefmark{3}}\\
\IEEEauthorblockA{\IEEEauthorrefmark{1}Visual Information Processing Laboratory,
Dept. of Computer Science \& Engg., IIT Kharagpur, India}\\
\IEEEauthorblockA{\IEEEauthorrefmark{2}XLIM UMR CNRS 7252, University of Poitiers, France}\\
\IEEEauthorblockA{\IEEEauthorrefmark{3}Dept. of Computer Science \& Engg., IIT Jodhpur, India}% <-this % stops an unwanted space
\vspace{-0.5cm}}
 %For a paper whose authors are all at the same %institution,
 %omit the following lines up until the closing ``}''.
 %Additional authors and addresses can be added with %``\and'',
 %just like the second author.
 %To save space, use either the email address or home %page, not both
%\and

\maketitle
%\ificcvfinal\thispagestyle{empty}\fi

%%%%%%%%% ABSTRACT
\begin{abstract}
   Deep neural networks have shown their profound impact on achieving human level performance in visual saliency prediction. However, it is still unclear how they learn the task and what it means in terms of understanding human visual system. In this work, we develop a technique to derive explainable saliency models from their corresponding deep neural architecture based saliency models by applying human perception theories and the conventional concepts of saliency. This technique helps us understand the learning pattern of the deep network at its intermediate layers through their activation maps. Initially, we consider two state-of-the-art deep saliency models, namely UNISAL and MSI-Net for our interpretation. We use a set of biologically plausible log-gabor filters for identifying and reconstructing the activation maps of them using our explainable saliency model. The final saliency map is generated using these reconstructed activation maps. We also build our own deep saliency model named cross-concatenated multi-scale residual block based network (CMRNet) for saliency prediction. Then, we evaluate and compare the performance of the explainable models derived from UNISAL, MSI-Net and CMRNet on three benchmark datasets with other state-of-the-art methods. Hence, we propose that this approach of explainability can be applied to any deep visual saliency model for interpretation which makes it a generic one.
\end{abstract}
%%%%%%%%% BODY TEXT
\section{Introduction}
Visual attention or selection is a significant perceptual function in human visual system (HVS) to process complex information from the complex natural scenes \cite{treisman1980feature}. The visual attention mechanism can extract essential features from redundant data to benefit the information processing in human brain. Bottom-up attention mechanisms are used to extract image features accompanied by visual information processing in human brain such as orientation, frequency, texture, color etc \cite{wang2005efficient}. As our nervous system has a limited ability to simultaneously process all the incoming sensory information, the attention selects and modulates the most relevant information. Multiple perceptive and cognitive operations, under a hierarchical control process, establish global priorities to highlight some locations, objects or features in the visual field \cite{wolfe2006sensation}. 
In recent times, researchers have collected free viewing style human gaze datasets of images and videos with the help of psycho-neurology experts. Much research effort has been expended on the development of selective visual attention models based on the saliency maps. The most efficient answer came out by employing deep neural networks (DNN), which in a way resemble the hierarchical structure of the processing in HVS \cite{wang2005efficient,barlow1961possible}. Provided that there exist a sufficient amount of saliency data, a DNN could learn the task of saliency detection to achieve a performance close to human perception of saliency. Yet, due to its black-box nature, it is inherently difficult to understand which aspects of the input data drive the decisions of the network. The impact of the kernel parameters of a DNN is visible on the activation maps at intermediate layers of a DNN. Hence, the contextual feature extraction process of visual data is encoded in those activation maps \cite{cerf2008predicting}. Understanding those maps and the corresponding kernel parameters of a deep saliency model using our knowledge of HVS help us in its interpretation and build an explainable deep saliency model.

In order to understand what a deep visual saliency model has learnt throughout its layers, we initially require to interpret its activation maps of intermediate layers \cite{yang2019dilated}. As we know, the kernel parameters learnt by the model during training are seen as N-D filters. Since it is difficult to exactly study and parameterize them, we instead examine their corresponding activation maps to decode a DNN. During information transmission, there involves a nonlinear association between activation maps in two consecutive layers of a DNN \cite{erdem2013visual}. A natural visual scene typically contains many objects of various structures at different scales. This suggests that the saliency detection should be carried out simultaneously at multiple different scales \cite{erdem2013visual}. On the other hand, this approach can explain the phonomena of primary visual cortex (V1) which produces multi-scale and multi-orientation features provided a stimulus \cite{li2020hvs}.

In terms of $HVS$, it is defined that the regions are salient if they differ from their surroundings \cite{hou2013visual}. Log-gabor or Difference of gaussian (DoG) filters are biologically plausible linear filters for this purpose. They are popular due to their resemblance to the receptive fields of neurons of HVS, namely the lateral geniculate nucleus of the thalamus (LGN) and V1 \cite{zipser1996contextual}. We follow the same approach as detailed in \cite{leboran2016dynamic} for generating the log-gabor filter responses. In the frequency domain, the transfer function of these filters is given as:
\begin{multline*}
    \Upsilon (\theta,\lambda) = \\ exp \bigg\{ -\frac{(log(\lambda_{0}/\lambda))^{2}}{2(log(\sigma_{f}/f_{0}))^{2}} \bigg\}.exp \bigg\{-\frac{(\theta-\theta_{0})^{2}}{2\sigma_{\theta}^2} \bigg\}
\end{multline*}
%\begin{multline*}
    %\Upsilon (r_{0},\theta_{0},\sigma_{r},\sigma_{\theta}) = \\ exp %\bigg\{ -\frac{(log(r/r_{0}))^{2}}{2(log(\sigma_{r}/r_{0}))^{2}} %\bigg\}.exp \bigg\{-\frac{(\theta-\theta_{0})^{2}}{2\sigma_{\theta}^{%2}} \bigg\}
%\end{multline*}
where $\theta$ and $\lambda$ represent the orientation and the wavelength of the filter respectively. And $\theta_{0}$, $\lambda_{0}$ (1/$f_{0}$), $\sigma_{f}$, and $\sigma_{\theta}$ are the center orientation, the center wavelength, the width parameter of the frequency, and the width parameter of orientation of the filter, respectively. Then, the generated filters are inverse transformed to the spatial domain and shifted to the center point. The filters in spatial domain are complex in nature. Thereby, the convolution of the filter with an image $I(x,y)$ results in a complex response $(O^{even}(x,y)+i.O^{odd}(x,y))$. The magnitude of that complex response represents the local energy at a particular 2D image coordinate of the image, represented by:
\begin{equation}
\label{eq:gaborenergy}
    E^{\theta,\lambda}(x,y)  = \sqrt{O^{even}(x,y)^{2}+O^{odd}(x,y)^{2}}
\end{equation}
Here, we consider the magnitude of the filter response since it is observed that the saliency has a deep connection with the variability in local energy of the filter responses \cite{leboran2016dynamic}.

In recent times, saliency methods \cite{adebayo2018sanity,chang2018explaining} have become a popular tool to explain the predictions of a trained model by highlighting parts of the input with some presumptions. However, most of them are focused on image classification networks and thus are not able to spatially differentiate between prediction explanations. They interpret the network decision process by producing explanation via bounding boxes \cite{karpathy2015deep} or attributes \cite{gulshad2019interpreting} providing textual justifications \cite{park2018multimodal} or generating low-level visual explanations \cite{zhou2016learning}. 

Apart from using saliency methods for explaining deep classification models, there are a good number of classical saliency models grounded in theories of preattentive vision such as the Feature Integration Theory \cite{treisman1980feature} or Guided Search Model \cite{wolfe2007guided} which mostly differ in their computational properties. Under the paradigm of biological plausibility, an early approach in \cite{itti1998model} was to add basic motion features to their static model. Later, the approach by \cite{cerf2008predicting} reformulated the previous model using concepts of graph theory. Over the time, there emerged at least three main groups of saliency models. In the first group, the models are developed by applying concepts drawn from Information theory and Natural Image Statistics. They use independent component analysis to learn basis functions from patches of a set of representative natural images. Representative examples of these models are AIM \cite{bruce2009saliency}, SUNDAY \cite{zhang2009sunday}, HCL \cite{hou2009dynamic}, etc. The second group of models follow Bayesian approaches for computation of saliency map by the evaluation of dissimilarities between the a priori and the a postriori distributions of the visual features around each point by using KL divergence \cite{cui2012temporal}. The third group includes compressed-domain visual saliency models which operate on the information found in a partial decoding of compressed video bitstream \cite{khatoonabadi2015compressed,petsiuk2018rise} and extract features such as block-based motion vectors, prediction residuals, block coding modes, etc. Even though there are a few attempts using perceptual theories to build a saliency model, they are only limited to classical approaches but not tried to explain the deep learning based architectures. Despite the superior performance of DNNs in computer vision tasks, DNNs are not still fully explainable.
%To our knowledge, there were also no attempts on parameterizing the kernels learnt by deep model with any of the known probabilistic distributions.

In this work, we develop a technique to derive explainable deep saliency models from their corresponding deep visual saliency models for visual saliency prediction. We do this by merging human perception theories and conventional concepts of visual saliency. This technique helps us understand the learning pattern of the deep network at its intermediate layers. For this purpose, we initially use two state-of-the-art deep saliency models namely UNISAL \cite{droste2020unified} and MSI-Net \cite{kroner2020contextual} for our interpretation which have MobileNetV2 \cite{sandler2018mobilenetv2} and VGG16 as the backbone networks, respectively. Since the information encoded by a deep network reflects in the pattern of activation maps, we would like to understand them at first. From human perception theories, we propose that the activation maps also resemble the responses of biologically plausible log-gabor filters. Visual saliency is perception-based and ground truth (GT) maps for datasets are built from humans’ gaze.
Since deep saliency models are trained with such datasets, DNN weights have similar distribution as log-gabor filters. Hence, we build a log-gabor filter bank whose parameters are chosen to ensure a broad coverage of the orientation space with uniform sampling and wavelengths of filters. We use this filter bank for interpreting and reconstructing the activation maps of UNISAL and MSI-Net. We employ the strategy of computing the block-wise variance with a block size of $8\times8$ for identifying the activation maps with respect to the block-wise variance computed filter responses. This is done because the second order statistical descriptors capture local structure information in an effective manner \cite{he2019understanding}. Also, we explain UNISAL and MSI-Net by imitating the $HVS$ and consider multiple scales of the activation maps for our processing. Eventually, we use these reconstructed activation maps to build the final saliency map.

Alongside, we propose a deep architecture for visual saliency prediction for images. Inspired by the image pyramid \cite{borji2019saliency} and skip layer architectures \cite{zeiler2014visualizing}, we build a feature extractor capturing the global and local contextual information at multiple scales from images. The features at different scales are concatenated along with local residual learning which we named Cross-concatenated Multi-scale Residual (CMR) block. A dilated inception module (DIM) \cite{yang2019dilated} is then used to enhance features with diverse field-of-views. While decoding, we up-sample features using convolutional layers to get back the input resolution. We also derive an explainable saliency model from CMRNet using the same technique as we did for UNISAL and MSI-Net. Then, we evaluate and compare the performance of the explainable models derived from UNISAL, MSI-Net and CMRNet on three benchmark datasets with the state-of-the-art models. To summarize, the main contributions of this work are multi-fold:
\begin{enumerate}
    \item A technique to derive the explainable classical saliency models from their corresponding deep neural architecture based saliency models using human perception theories,
    \item Near approximation of the statistics of inner layer activation maps with a set of biologically plausible log-gabor filter responses,
    \item Second order statistical descriptors for identifying the activation maps with respect to log-gabor filter responses, and
    \item Explainable saliency models are derived from UNISAL, MSI-Net and a newly proposed deep saliency architecture (CMRNet).
\end{enumerate}

\section{Explainable Deep Saliency Model}
By understanding the above mentioned phenomena, we build an explainable saliency model for predicting visual saliency based on the existing theories of human perception. We derive this model from UNISAL where the latter has approximately $75$ deep layers in it. We approximate the kernel parameters of those layers with the generated log-gabor filter bank. We know that the activation maps of a layer of UNISAL are combined in a certain way to give out the activation maps of the very next layer. Hence, in our explainable saliency model, we try to generate the activation maps of deep layers using the filter bank. Then, we compute the Mean Absolute Error (MAE) as a similarity metric of activation maps with the log-gabor filter responses. The inverse of the least $10$ MAE values of an activation map with filter responses are taken because lower the MAE, higher the contribution of that filter for reconstruction. Since reconstruction happens at every layer, we have empirically chosen top 10 filter responses for limiting the number of computations. Then, those inverted MAEs are unit normalized so that their sum equals to $1$. These are used as coefficients for linear combination to reconstruct the activation maps at a particular layer. Those reconstructed maps are again passed through the log-gabor filter bank and considered for computations in the next layer. This process continues in parallel to the UNISAL until we reach the last layer of it, where we end up getting the final saliency map. In the same way, we derive the explainable deep saliency model from MSI-Net also, which has approximately $30$ deep layers in it.

\subsection{Intermediate Activation Maps of UNISAL}
\begin{figure}[h]
\centering
\includegraphics[width=0.9\columnwidth]{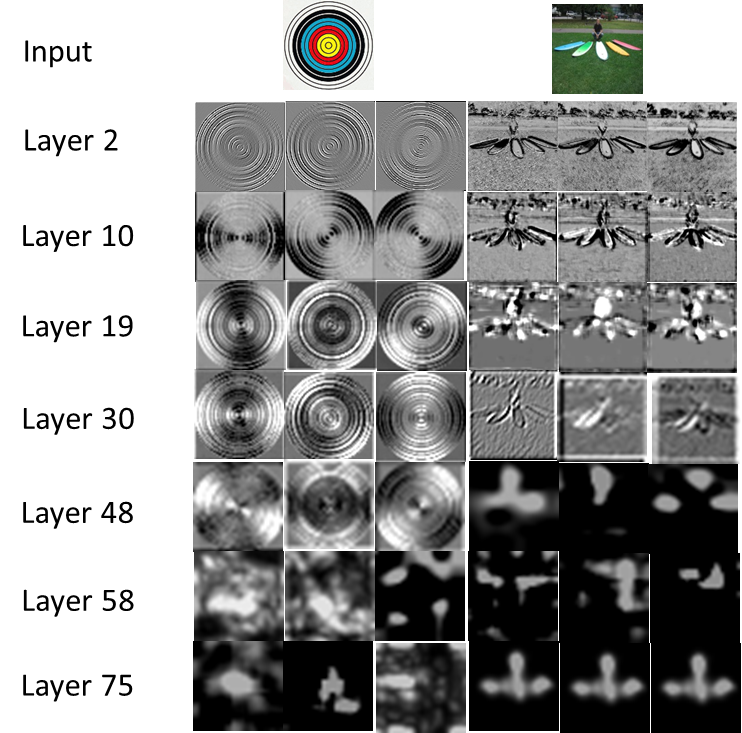}
\caption{Activation maps from various layers of UNISAL.}
\label{fig:actmapsunisal}
\end{figure}

In Fig.\ref{fig:actmapsunisal}, we show some sample activation maps from intermediate layers of UNISAL. \textit{Layer 2} corresponds to the one just after the input layer whose maps have highlighted oriented edges. The extraction of edges is identified to capture the low level information which is also considered as an intermediate saliency map. This is because the gradients carry prominent information during the process of learning. In $HVS$, the neurons codify sensory information by reducing the redundancy along the visual pathway \cite{barlow1961possible}.

It is interesting to observe that the oriented edges are profoundly visible as move little deeper (towards \textit{Layer 19}). By observing the maps after \textit{Layer 30} till \textit{Layer 48}, we understand that the model starts to learn extracting the salient regions. The maps are found to have regions beings identified rather than the edges as detected in earlier layers. The oriented edges gradually become denser and region selective as the model starts to learn the task. From \textit{Layer 50}, the UNISAL gets indulged in efficiently extracting the salient regions. This is expected since in any deep learning model, the top layers work on extracting low level features like contrast, orientation, etc, whereas the deeper layers extract the abstract and contextual information related to the task being given \cite{he2019understanding}.

\subsection{Methodology}
\label{sec:methodology}
In this work, we propose to explain the functioning of UNISAL by building an explainable saliency model. We devise a methodology based on some of the psychovisual perceptual theories. The block diagram in Fig. \ref{fig:mainblock} shows that approach for building an explainable saliency model by near parameterizing the UNISAL kernels using a log-gabor filer bank. We consider UNISAL which is trained on both the image and video saliency datasets (SALICON \cite{jiang2015salicon}, DHF1K \cite{wang2018revisiting}, Hollywood-2 \& UCF-sports \cite{mathe2014actions}) for our purpose.
\subsubsection{Log-gabor Filter Bank}
\label{sec:loggabor}
Here, our aim is to produce a filter bank that provides even coverage of the section of the spectrum we wish to represent. Also, we ensure that the outputs of individual filters in the bank are as independent as possible. This results in filters being arranged efficiently to provide as much information as possible. Thus, the transfer functions of the filters need to have the minimum overlap to achieve fairly even spectral coverage.

In this work, we uniform sample the orientation space into $5$ regions and $4$ different wavelengths. The first filter of each orientation has a wavelength ($\lambda_{0}$) of $1$ while the scaling factor between successive filters is $1.5$. This results in $20$ log-gabor filters for varying $\theta$ and $\lambda$. We also vary both $\sigma_{f}$ and $\sigma_{\theta}$ with initial values of $0.5$ and scaling up by a factor of $1.5$ simultaneously for $4$ times. This results in $4$ filters by varying $\sigma_{f}$ and $\sigma_{\theta}$. Hence, there are $20$ filters by varying $\lambda$, $\theta$ and $4$ filters by varying $\sigma_{f}$, $\sigma_{\theta}$. Therefore, as a whole, we build a set of $80$ log-gabor filters for our analysis.
\subsubsection{Activation Maps of Intermediate Layers at Different Scales}
As we know, the first layer of any deep architecture would be the input layer, which processes the provided input image as it is. Therefore, we move on to the activation maps at \textit{Layer 2}. This layer is just after the input layer in UNISAL which has A activation maps in it. In order to understand the extracted local features (as activation maps) efficiently, we consider $5$ different scales of these maps. 
%We found it out by an ablation study with $5$ different cases in terms of number of scales: 1) original, 2) 3, 3) 5, 4) 7, and 5) 9 different scales. 
Each of those A maps are upscaled and downscaled twice individually by a factor of $2$. This results in two different scales of maps. Along with them, we also consider the original scale too, thus making them $5$ different scales totally. Hence, we represent the activation maps by $Act^{n}_{a,s}$ at $n^{th}$ layer, where $n=1..N$ layers, $a=1..A$ activation maps and $s=1..S$ scales..

\begin{figure*}[h]
\centering
\includegraphics[width=0.8\textwidth]{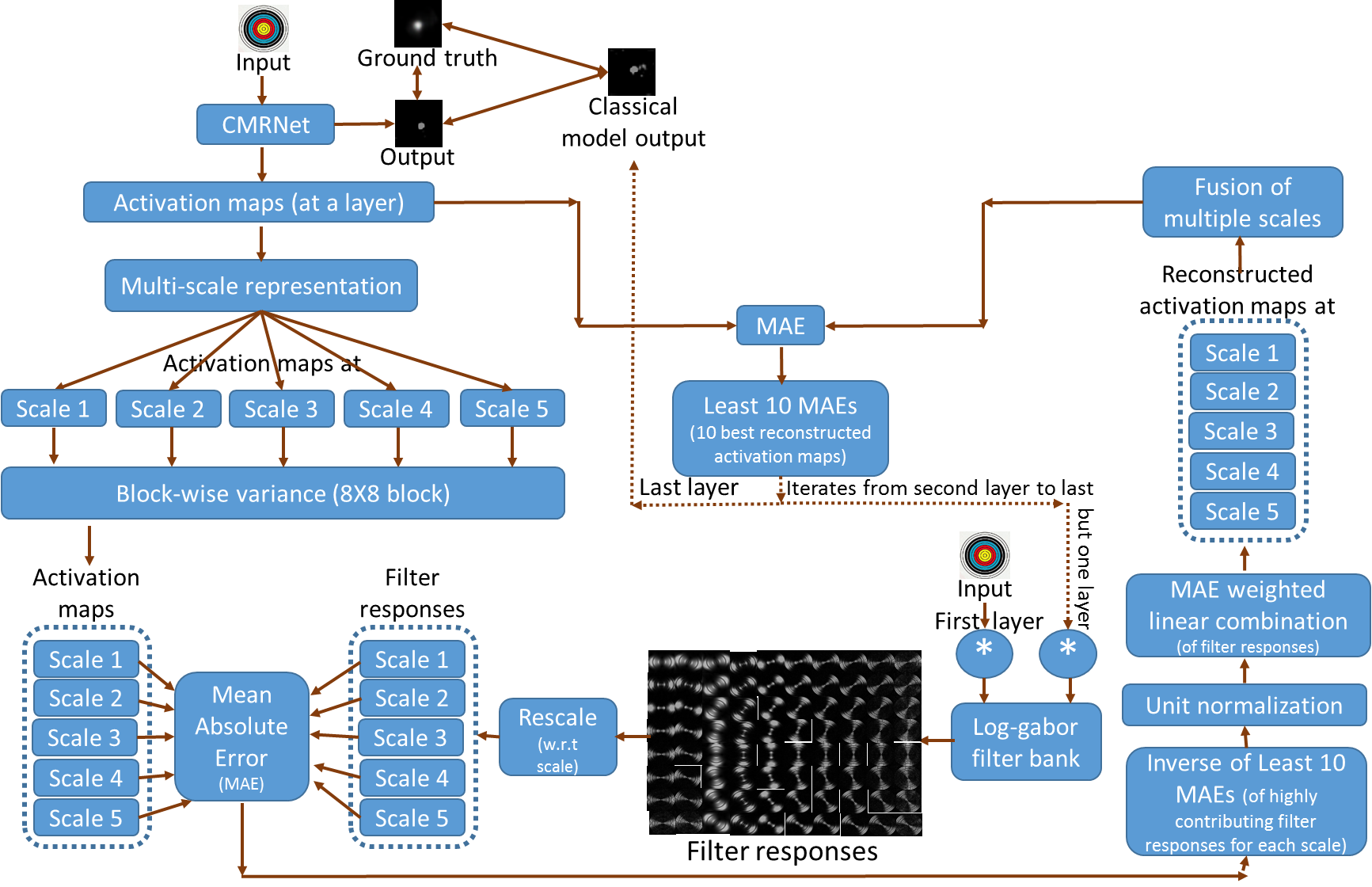}
\caption{Sequence of processes for deriving explainable deep saliency models.}
\label{fig:mainblock}
\end{figure*}

As explained in Section \ref{sec:loggabor}, we consider the generated log-gabor filter bank for our processing. For implementing our proposed approach on activation maps at \textit{Layer 2}, we generate the responses of the filter bank by convolving the input image with all of them. Then, we compute the local energy at each pixel location from the real and imaginary parts of the response as given in Eq. \ref{eq:gaborenergy}. This is because the kernels of \textit{Layer 2} work on the input image directly. We get $80$ filter responses as our initial set of responses and we use them to reconstruct the activation maps of \textit{Layer 2} and represent those responses by $F_{\theta,\lambda}$.

For the ease of understanding, we consider in detail the approach at \textit{Layer 2}, where $n=2$ at a particular scale ($s$). The generated filter responses may not be of the same spatial resolution as the activation maps. Therefore, we resize the filter responses to the scale of that particular activation map. We then compute the block-wise variance with a block size of $8\times8$ for both the activation map and the filter responses. The computed variance of a block $b$ for an activation map $Act^{n}_{a,s}$ is represented by $V^{b}_{Act}$, where $b=1..B$ blocks. Similarly, it is represented by $V^{b}_{F}$ for a log-gabor response $F_{\theta,\lambda}$. Then, the mean absolute error between the variance evaluated activation map and the filter response is 
\begin{equation}
    MAE(V_{Act},V_{F}) = \frac{\sum_{b=1}^{B} \mid V^{b}_{Act}-V^{b}_{F} \mid}{B}
\end{equation}

By comparing all the A activation maps with $80$ filter responses for each of the $5$ scales, we represent those MAEs by $MAE^{n}_{a,g,s}(V_{Act},V_{F})$, where $g=1..80$ responses, $a=1..A$ activation maps, and $s=1..5$ scales. We know that the MAE is a dissimilarity metric and the lower the MAE, the higher the similarity between the activation map and the filter response. In order to reconstruct an activation map, we choose the least $10$ MAEs along with their filter responses. Those $10$ responses are found to be highly contributing for reconstructing that activation map. We represent those $10$ MAE values and the corresponding filter responses by $MAE^{n}_{a,h,s}$ and $F^{n}_{h,s}$, respectively for $h=1..10$ at scale $s$ for $a^{th}$ activation map. 

Having found the top $10$ highly contributing filter responses and their corresponding MAEs, we invert them because highly contributing filters show lower MAE values. Then, we perform unit normalization on those $10$ inverse MAEs so that their sum equals to $1$. They are used as coefficients during linear combination for reconstruction of maps. This is represented by
\begin{equation}
    R_{Act^{n}_{a,s}} = \sum_{h=1}^{10} \frac{ \frac{1}{MAE^{n}_{a,h,s}}}{\sum_{h=1}^{10} \frac{1}{MAE^{n}_{a,h,s}}} .F^{MAE^{n}}_{h,s}, s=1..5
\end{equation}
where $MAE^{n}_{a,h,s}$ and $F^{MAE^{n}}_{h,s}$ are those $10$ MAEs and their corresponding log-gabor filter responses, respectively.
\subsubsection{Fusion of Activation Maps at Multiple Scales}
Since the reconstructed activation maps at different scales contribute to encode various structures at different resolutions, we need to fuse them for efficient reconstruction. We rescale the reconstructed maps to the scale of the original activation map at first. Then, we build the final activation map as the sum of the squared root of those at different scales. It is computed as
\begin{equation}
    R_{Act^{n}_{a}} = \sqrt{ \sum_{s=1}^{S} R_{Act^{n}_{a,s}}^{2} }, a=1..A
\end{equation}
After reconstruction, it is necessary to evaluate how far we are able to effectively do that because we use them for processing in the next layer. Therefore, we evaluate the MAE of the reconstructed activation maps with their corresponding original activation maps as 
\begin{multline*}
    MAE^{n}_{a}(Act^{n}_{a},R_{Act^{n}_{a}}) = \\ \frac{\sum_{x=1}^{H} \sum_{y=1}^{W} \mid Act^{n}_{a}(x,y)-R_{Act^{n}_{a}}(x,y) \mid}{H\times W}, a=1..A
\end{multline*}
As discussed earlier, lower the MAE, higher the similarity between two images. Hence, we choose the best $10$ reconstructed activation maps with least $10$ MAE values for further processing as shown in Fig. \ref{fig:mainblock}. These other $10$ maps are selected corresponding to different activation maps. We perform the above mentioned set of operations iteratively at all the layers to realize their activation maps. Then, we evaluate the performance of the explainable saliency model derived from UNISAL quantitatively and quantitatively on three benchmark datasets. In the same way, we also derive this explainable model on MSI-Net also. Further, we apply our proposed approach of explaining a deep saliency model on the newly proposed deep saliency architecture CMRNet. 
\section{CMRNet Architecture}
\begin{figure}[h]
\centering
\subfloat[CMRNet architecture]{\includegraphics[width=0.48\columnwidth]{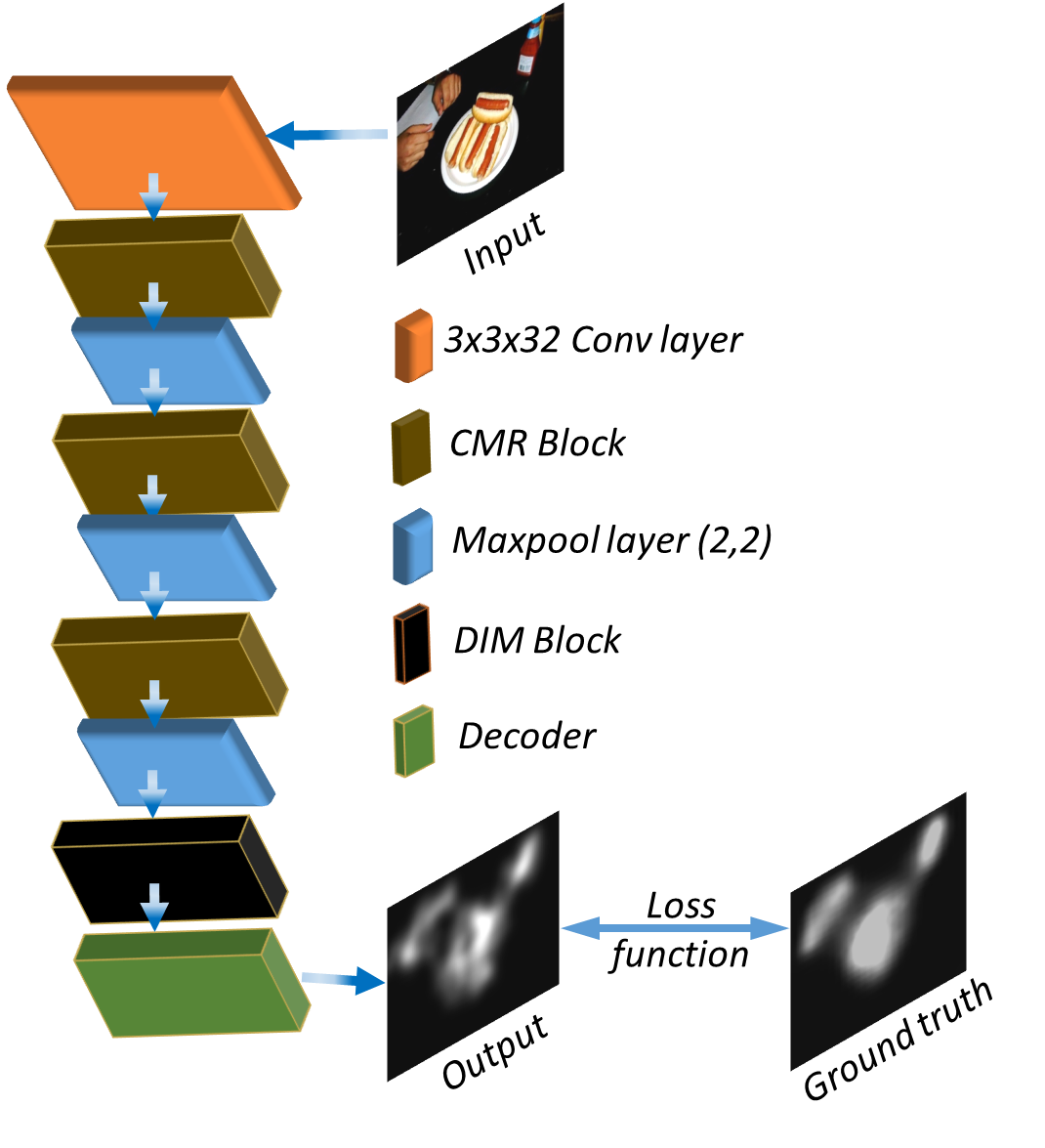}\label{fig:cmrnet}}
\hfill
\subfloat[CMR Block]{\includegraphics[width=0.52\columnwidth]{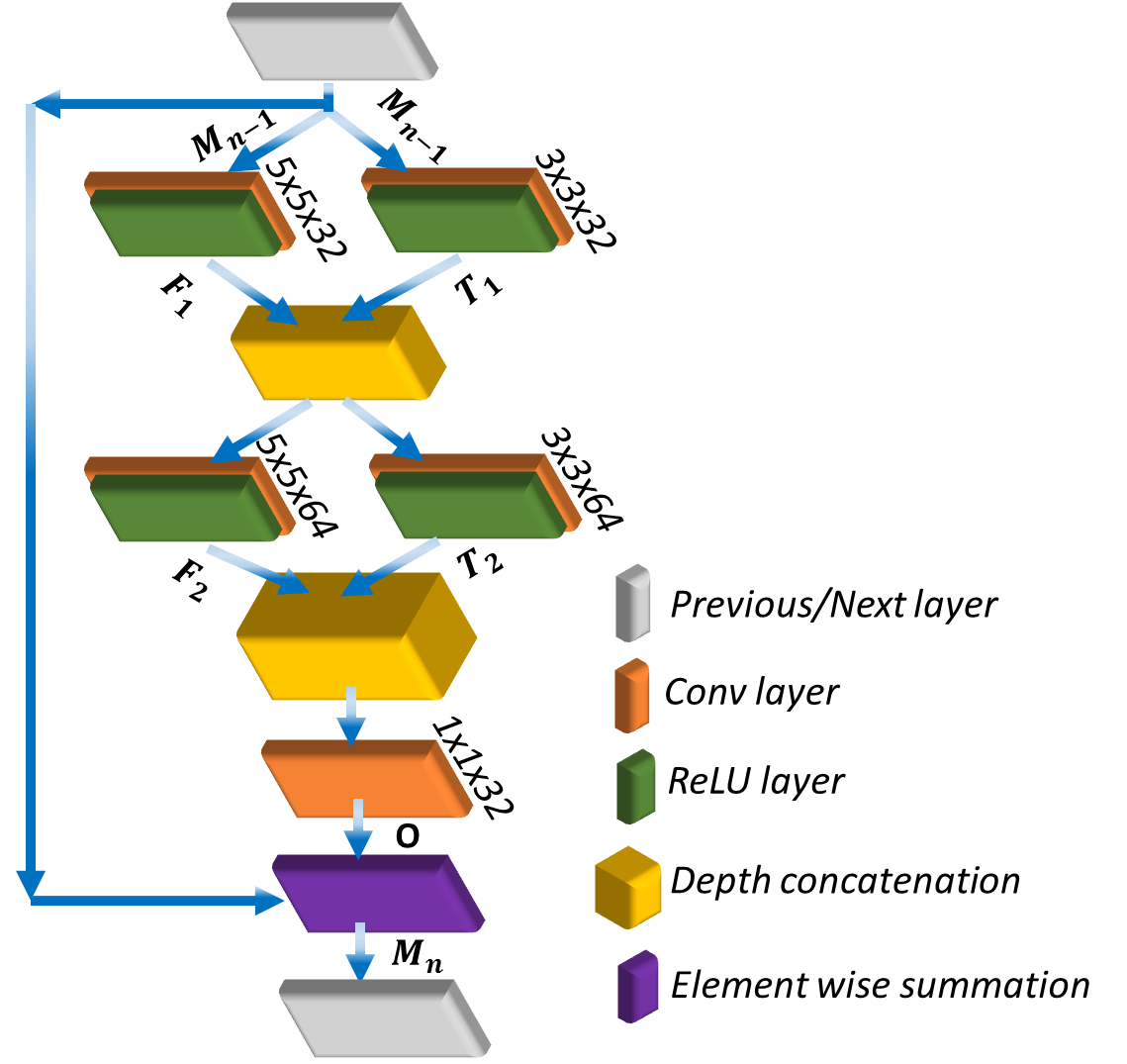}\label{fig:cmrblock}}
\caption{CMRNet architecture and CMR block}
\label{fig:cmrnetandblock}
\end{figure}

We develop an architecture named Cross-concatenated Multi-scale Residual block based network (CMRNet) for visual saliency prediction as shown in Fig. \ref{fig:cmrnet}. We build CMRNet with three vital blocks: 1) A Cross-concatenated Multi-scale Residual (CMR) block, 2) A Dilated Inception Module (DIM) \cite{yang2019dilated}, and 3) A Decoder. We propose a multi-scale feature extractor with added residual part (CMR block) in it to capture both the global and local multi-scale information. A series of three CMR blocks are used with max pooling in between them, to lessen the number of computations. Then, we propose to attach a DIM on top of them. This DIM block diversifies the receptive fields of the extracted features from CMR blocks. Then, they are followed by a series of convolutional layers, a sigmoid layer, and a bi-cubic up-sampling layer which together act as a decoder. We do not use any deconvolutional layers in the decoder because they need heavy computations and also result in non-smoothing operations inside them \cite{odena2016deconvolution}. 

\subsection{Cross-concatenated Multi-scale Residual (CMR) block}
\textbf{Cross-concatenation of Multi-scale Features:} We build a two-path network (Fig. \ref{fig:cmrblock}) where these paths use different sized filters. The interaction between those paths help in better feature extraction, where those operations are formulated as
\begin{equation}\label{eq:1}
    T_{1} = \sigma(w_{3 \times 3}^{1}*M_{n-1}+b^{1}),
\end{equation}
\begin{equation}\label{eq:2}
    F_{1} = \sigma(w_{5 \times 5}^{1}*M_{n-1}+b^{1}),
\end{equation}
\begin{equation}\label{eq:3}
    T_{2} = \sigma(w_{3 \times 3}^{2}*[T_{1},F_{1}]+b^{2}),
\end{equation}
\begin{equation}\label{eq:4}
    F_{2} = \sigma(w_{5 \times 5}^{2}*[T_{1},F_{1}]+b^{2}),
\end{equation}
\begin{equation}\label{eq:5}
    O^{'} = w_{1 \times 1}^{3}*[T_{2},F_{2}]+b^{3},
\end{equation}
where $w$ and $b$ represent the weight and bias, respectively. The superscripts represent the layer number at which they are placed while the subscripts represent the convolutional filter size at that particular layer, and $\sigma(.)$ is used to show the ReLU function. Also, the depth concatenation of two feature maps $T_{1}$ and $F_{1}$ are shown as $[T_{1},F_{1}]$. 

\begin{figure}[h]
\centering
\subfloat[DIM block]{\includegraphics[width=0.6\columnwidth]{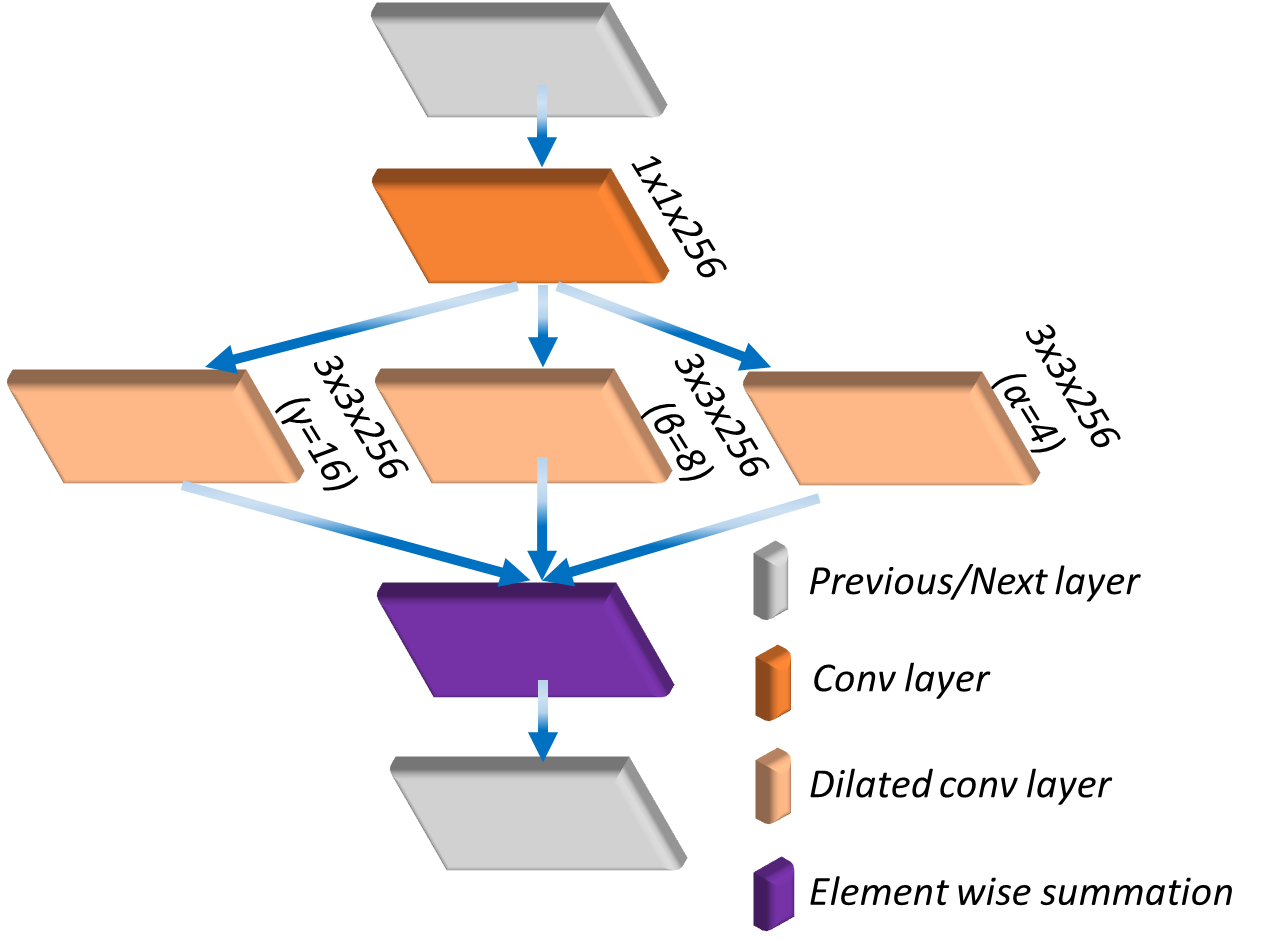}\label{fig:dim}}
\hfill
\subfloat[Decoder]{\includegraphics[width=0.4\columnwidth]{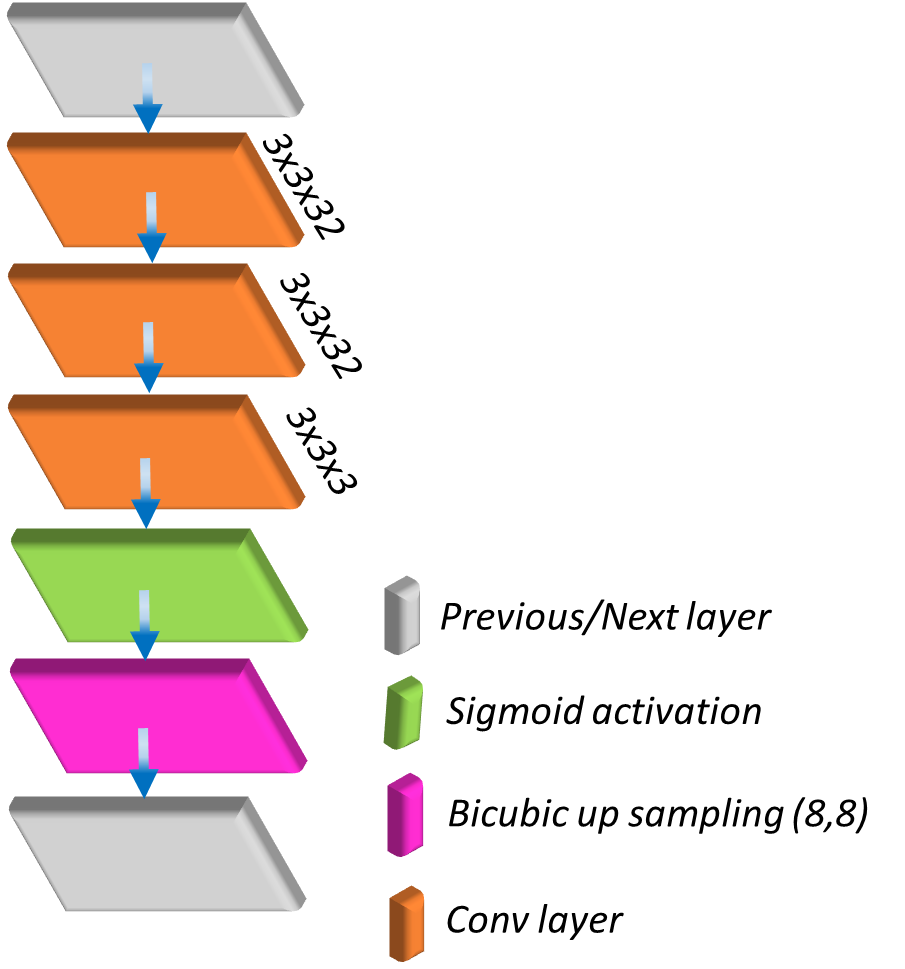}\label{fig:decoder}}
\caption{Dilated inception module and the decoder}
\label{fig:dimanddecoder}
\end{figure}

\textbf{Local Residual Learning:} We adapt residual learning to each CMR block for better efficiency and represented as:
\begin{equation}\label{eq:6}
    M_{n} = O^{'} \oplus M_{n-1},
\end{equation}
where $M_{n}$ and $M_{n-1}$ are the input and output of the $CMR$ block, respectively. $O^{'} \oplus M_{n-1}$ is the element-wise addition. We find the local residual learning reduces the computational complexity and improves the performance. In our experiments, we find that $[\alpha, \beta, \gamma]$ as $[4, 8, 16]$ shows the best performance for DIM. In decoder (Fig. \ref{fig:decoder}), we use a bi-cubic up-sampling layer to make the output resolution similar to the input and to reduce the computation cost.

\textbf{Loss Function:} We build a loss function to make sure that saliency maps are invariant to their maximum value. Hence, we modify the $l$1-norm with linearly normalized pixel values as
\begin{equation}\label{eq:lossfunction}
    L(p,g) = \sum_{i} |p_{i}-g_{i}|
\end{equation}
where $p$, $g$ are the predicted and ground truth saliency maps, respectively, with $p_{i}$ and $g_{i}$ given as
\begin{equation}\label{eq:pigiequation}
    p_{i} = \frac{x_{i}^{p}}{\sum_{i=1}^{N}x_{i}^{p}}, g_{i} = \frac{x_{i}^{g}}{\sum_{i=1}^{N}x_{i}^{g}}
\end{equation}
where $x_{i}$ is the unnormalized pixel value for either $p$ or $g$.

\subsection{Experimental Evaluation}
We use the following three datasets for our analysis:
SALICON contains $20,000$ images taken from the Microsoft COCO dataset \cite{lin2014microsoft}. Out of them, $10,000$ are for training, $5000$ for validation, and $5000$ for testing. MIT1003 \cite{judd2009learning} contains $1003$ random images taken from Flickr and LabelMe. MIT300 \cite{judd2012benchmark} contains $300$ natural images from both indoor and outdoor scenes.

We do not use any transfer learning procedure, hence we train CMRNet from scratch. It is trained using Adam optimizer \cite{kingma2014adam} with an initial learning rate of $10^{-3}$ which is scaled down by a factor of $0.1$ after every two epochs. Our model is trained for $60$ epochs with a batch size of $8$. The training is engaged using SALICON training set and its validation set is used to validate the model. We evaluate the trained model on both MIT1003 and MIT300 datasets, for which we resize their input images to $320\times480$. We consider Similarity (SIM) \cite{judd2012benchmark}, Linear Correlation Coefficient (CC) \cite{cornia2018predicting}, AUC shuffled (sAUC) \cite{cornia2018predicting}, and AUC Judd \cite{kruthiventi2017deepfix} as the evaluation metrics. Hereafter, we consider this trained CMRNet for building an explainable saliency model.

\begin{figure}[h]
\centering
\includegraphics[width=0.9\columnwidth]{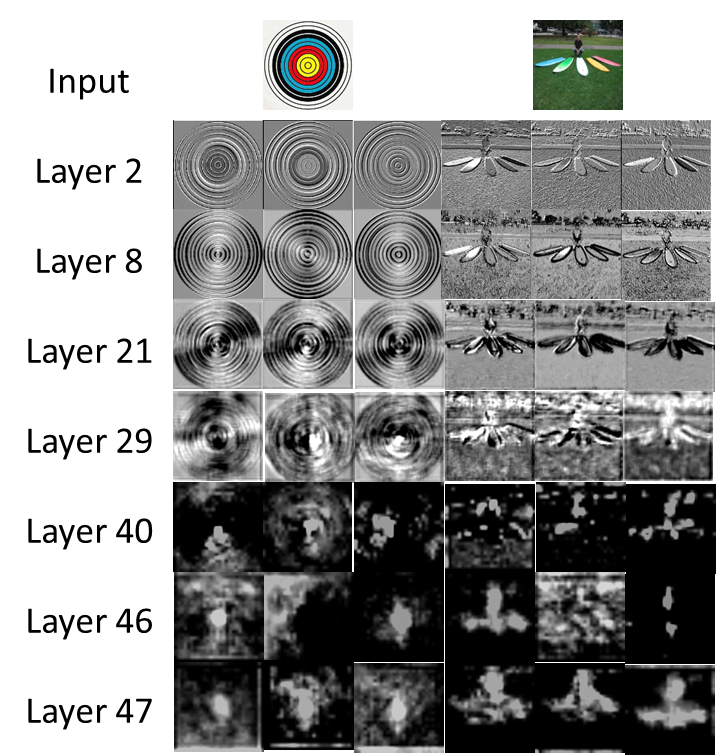}
\caption{Activation maps from various layers of CMRNet.}
\label{fig:actmaps}
\end{figure}

Fig. \ref{fig:actmaps} shows some sample activation maps from intermediate layers of CMRNet. Just like in Fig. \ref{fig:actmapsunisal} where we have shown activation maps of UNISAL, we find that the initial layers (\textit{Layer 2} till \textit{Layer 21}) highlight the low-level features like oriented edges, whereas the inner layers extract the contextual information related to the given task.

We use the same technique detailed in Section \ref{sec:methodology} for deriving an explainable classical saliency model from CMRNet. At the end, we compare its performance with other state-of-the-art models on three benchmark datasets.

\section{Results and Discussion}

\subsection{Quantitative analysis}
We evaluate the performance of the explainable saliency models derived from UNISAL, MSI-Net and CMRNet on SALICON validation, MIT1003 and MIT300 benchmark datasets as shown in Tables \ref{table:salvalqualitative}, \ref{table:mit1003qualitative}, and \ref{table:mit300qualitative}, respectively. The second column represents if the model is a DNN ($Y$) or a classical model ($N$). The proposed explainable model is not self sustained, rather it is a generic approach to interpret any
deep saliency model. The comparison with classical models is shown to demonstrate the effectiveness of the proposed approach in explaining a DNN in terms of principles behind the classical models. Among the classical methods, the best performing model is shown in bold and italicized font. While among the DNN models, it is shown in bold font alone.

From Tables \ref{table:salvalqualitative} and \ref{table:mit300qualitative} (among the classical models), the explainable saliency model derived from UNISAL performs at par or even the best with the state-of-the-art models. This is expected since in deep models, UNISAL is found to be the best one which surely would reflect in its explainable saliency model too. However, the explainable model from CMRNet performs a bit inferior to the other classical methods as the latter are built by understanding their exact predecessor classical model and modify accordingly for even better performance. However, our explainable model is derived from a deep model and built with an intention to interpret it rather than improving an existing classical saliency model. The explainable model derived from UNISAL is found to slightly outperform the one derived from CMRNet, as UNISAL network itself is found performing better than CMRNet. Since UNISAL has a pre-trained backbone (MobileNetV2) network unlike our CMRNet, it mayperform better than CMRNet. However, from Table \ref{table:mit1003qualitative}, we see that MSI-Net performs better than UNISAL with a very marginal difference, and seen as very close in performance. Hence, we see that the performance of explainable model derived from CMRNet is comparatively closer to the best. It is also found to outperform ITTI and SIMgrouping models in some metrics. We observe that the closeness of the metrics for explainable models from CMRNet, UNISAL and MSI-Net indicate their similarity in performance.

From Tables \ref{table:salvalqualitative} and \ref{table:mit300qualitative} (among the DNN models), UNISAL is found to perform the best out of the existing methods. And, CMRNet performs at par with other state-of-the-art models, but may not outperform them. From Table \ref{table:comprecomp}, we observe that every other model is heavier than CMRNet model by at least $6.2$ times in terms of number of model parameters. UNISAL with backbone as MobileNet has $15.5$ million parameters, lies in the closest vicinity. Also, the average inference time is $0.02$ sec per image which outperforms all the other models. This happens because our network neither adapts any transfer learning nor depends on the existing huge architectures. Models with backbone networks like VGG16 and MobileNet are huge in parameters and take more time for data transmission through the network. However, our CMRNet is significantly smaller in size and faster during inference.

On SALICON validation set (Table \ref{table:salvalqualitative}), we find that our CMRNet varies by $1.7\%$ SIM, $1.7\%$ CC, $2.9\%$ sAUC, $1.2\%$ AUC Judd compared with the best. On MIT1003 (Table \ref{table:mit1003qualitative}), it varies by $2.8\%$ SIM, $6.8\%$ CC, $4.9\%$ sAUC, $2.5\%$ AUC Judd with the best. Similarly, on MIT300 (Table \ref{table:mit300qualitative}), it varies by $4.2\%$ SIM, $3.75\%$ CC, $2.7\%$ sAUC, $2.3\%$ AUC Judd with the best. From Table \ref{table:mit1003qualitative}, we see that MSI-Net performs better than UNISAL with a very marginal difference.

Since the UNISAL (Tables \ref{table:salvalqualitative} and \ref{table:mit300qualitative}) or MSI-Net (Table \ref{table:mit1003qualitative}) outperforms our CMRNet, it is even meaningful to see that there is a correspondence in their performances of their explainable saliency models. This shows that our approach for explainability goes with the performance and learning capabilities of the corresponding deep model. Comparing the classical models with their corresponding deep networks, we observe that the latter is superior in performance. The very reason being that the explainable model may not fully interpret the intricacies of a deep architecture. Over the past few years, though there are a very few attempts to unlock and explore deep architectures, it is still a partly answered question.

\begin{table}[h]
\centering
\caption{Performance comparison of our derived explainable saliency models with other classical models for SALICON validation set.}
\begin{adjustbox}{max width=\columnwidth}
\begin{tabular}{c|c|c|c|c|c}
\toprule
\textbf{Methods} & \textbf{DNN} & \textbf{SIM}   & \textbf{CC}    & \textbf{sAUC}  & \textbf{AUC Judd} \\ \midrule
\textbf{ITTI \cite{itti1998model}} & N            & 0.495          & 0.636          & 0.710          & 0.744             \\ %\hline
\textbf{ICL \cite{hou2009dynamic}} & N             & \textbf{\textit{0.553}} & 0.697 & 0.72           & 0.839             \\ %\hline
\textbf{HFT \cite{li2012visual}}  & N            & 0.549          & 0.676          & 0.722          & \textbf{\textit{0.867}}    \\ %\hline
\textbf{SIMgrouping \cite{parragalow}} & N     & 0.441          & 0.656          & \textbf{\textit{0.771}} & 0.794             \\ %\hline
\textbf{RARE \cite{riche2012rare}} & N            & 0.513          & \textbf{\textit{0.731}}          & 0.768          & 0.847             \\ %\hline
\textbf{Explainable CMRNet}  & N           & 0.533           & 0.7           & 0.76           & 0.804             \\ %\hline
\textbf{Explainable UNISAL}  & N           & 0.549           & 0.720           & \textbf{\textit{0.771}}           & 0.815             \\ %\hline
\textbf{Explainable MSI-Net}  & N           & 0.54           & 0.705           & 0.769           & 0.804             \\ \midrule
\textbf{CMRNet (Ours)} & Y          & 0.728          & 0.846          & 0.788          & 0.886             \\ %\hline
\textbf{UNISAL \cite{droste2020unified}} & Y      & \textbf{0.741}          & \textbf{0.861}          & \textbf{0.812} & \textbf{0.897}                                                                             \\ %\hline
\textbf{MSI-Net \cite{kroner2020contextual}} & Y      & 0.732          & 0.851          & 0.797 & 0.890                                                                             \\ \bottomrule
\end{tabular}
\end{adjustbox}
\label{table:salvalqualitative}
\end{table}

\begin{table}[h]
\centering
\caption{Performance comparison of our explainable saliency models with other classical models for MIT1003 dataset.}
\begin{adjustbox}{max width=\columnwidth}
\begin{tabular}{c|c|c|c|c|c}
\toprule
\textbf{Methods} & \textbf{DNN} & \textbf{SIM}   & \textbf{CC}    & \textbf{sAUC}  & \textbf{AUC Judd} \\ \midrule
\textbf{ITTI \cite{itti1998model}} & N            & 0.362          & 0.576      & 0.610          & 0.674             \\ %\hline
\textbf{ICL \cite{hou2009dynamic}}  & N            & 0.420 & 0.637 & 0.617          & 0.769             \\ %\hline
\textbf{HFT \cite{li2012visual}}  & N             & 0.416          & 0.616          & 0.619          & \textbf{\textit{0.797}}    \\ %\hline
\textbf{SIMgrouping \cite{parragalow}} & N      & 0.308          & 0.596          & 0.668 & 0.724             \\ %\hline
\textbf{RARE \cite{riche2012rare}}  & N            & 0.380          & 0.617          & 0.665          & 0.777             \\ %\hline
\textbf{Explainable CMRNet} & N             & 0.415          & 0.64         & 0.653          & 0.738             \\ %\hline
\textbf{Explainable UNISAL} & N             & 0.426          & 0.668         & 0.673          & 0.748             \\ %\hline
\textbf{Explainable MSI-Net} & N             & \textit{\textbf{0.435}}          & \textit{\textbf{0.70}}         & \textit{\textbf{0.693}}          & 0.761             \\ \midrule
\textbf{CMRNet (Ours)}           & Y            & 0.68           & 0.81           & 0.78           & 0.887             \\ %\hline
\textbf{UNISAL \cite{droste2020unified}}           & Y            & 0.691           & 0.85           & 0.79           & 0.897             \\ %\hline
\textbf{MSI-Net \cite{kroner2020contextual}}          & Y            & \textbf{0.70}           & \textbf{0.87}           & \textbf{0.81}           & \textbf{0.91}              \\ \bottomrule
\end{tabular}
\end{adjustbox}
\label{table:mit1003qualitative}
\end{table}

\begin{table}[h]
\centering
\caption{Performance comparison of our explainable saliency models with other classical models for MIT300 dataset.}
\begin{adjustbox}{max width=\columnwidth}
\begin{tabular}{c|c|c|c|c|c}
\toprule
\textbf{Methods}      & \textbf{DNN} & \textbf{SIM}            & \textbf{CC}            & \textbf{sAUC}          & \textbf{AUC Judd}       \\ \midrule
\textbf{ITTI \cite{itti1998model}}         & N            & 0.405                   & 0.495                  & 0.55                   & 0.70                    \\ %\hline
\textbf{ICL \cite{hou2009dynamic}}        & N            & 0.472                   & 0.562                  & 0.6                    & 0.75                    \\ %\hline
\textbf{HFT \cite{li2012visual}} & N            & 0.319                   & 0.379                  & 0.502                  & 0.652                   \\ %\hline
\textbf{SIMgrouping \cite{parragalow}}     & N            & 0.343                   & 0.433                  & 0.504                  & 0.654                   \\ %\hline
\textbf{RARE \cite{riche2012rare}}         & N            & 0.506 & 0.595                  & 0.64 & 0.806 \\ %\hline
\textbf{Explainable CMRNet}         & N            & 0.504                   & 0.64 & 0.635                  & 0.804                   \\ %\hline
\textbf{Explainable UNISAL}         & N            & \textit{\textbf{0.524}}                   & \textit{\textbf{0.70}} & \textit{\textbf{0.675}}                  & \textit{\textbf{0.824}}                   \\ %\hline
\textbf{Explainable MSI-Net}         & N            & 0.515                   & 0.683 & 0.668                  & 0.814                   \\ \midrule
\textbf{CMRNet (Ours)}       & Y            & 0.68                    & 0.77                   & 0.70                   & 0.85                    \\ %\hline
\textbf{UNISAL \cite{droste2020unified}}      & Y            & \textbf{0.71}           & \textbf{0.80}          & \textbf{0.72}          & \textbf{0.87}           \\ %\hline
\textbf{MSI-Net \cite{kroner2020contextual}}      & Y            & 0.691           & 0.787          & 0.716          & 0.863           \\ \bottomrule
\end{tabular}
\end{adjustbox}
\label{table:mit300qualitative}
\end{table}

\begin{table}[h]
\centering
\caption{Comprehensive comparison with state-of-the-art models on SALICON validation set}
\begin{adjustbox}{max width=\columnwidth}
\begin{tabular}{c|c|c}
\toprule
\textbf{Model}           & \textbf{\begin{tabular}[c]{@{}c@{}}\#parameters\\ ($\times10^{6}$)\end{tabular}} & \textbf{\begin{tabular}[c]{@{}c@{}}Average inference time\\ (per image, in sec)\end{tabular}} \\ %\hline
\midrule
\textbf{CMRNet (Ours)}  & \textbf{2.5}                     & \textbf{0.02}                                                                       \\ %\hline
\textbf{DINet \cite{yang2019dilated}}               & 27.04          & 0.06                                                                                  \\ %\hline
\textbf{SAM-ResNet \cite{cornia2018predicting}}               & 70.09                                                                    & 0.09                                                                                  \\ %\hline
\textbf{DSCLRCN \cite{liu2018deep}}      & \textgreater{}33.71                                                           & 0.27                                                                                  \\ %\hline
\textbf{DVA \cite{wang2017deep}}                        & 25.07                                                                          & 0.03                                                                                  \\ %\hline
\textbf{UNISAL \cite{droste2020unified}}                            & 15.5                                                               & 0.028                                                                                  \\ %\hline
\bottomrule
\end{tabular}
\end{adjustbox}
\label{table:comprecomp}
\end{table}

\begin{figure*}[h]
\centering
\subfloat[CMRNet]{\includegraphics[width=0.5\textwidth]{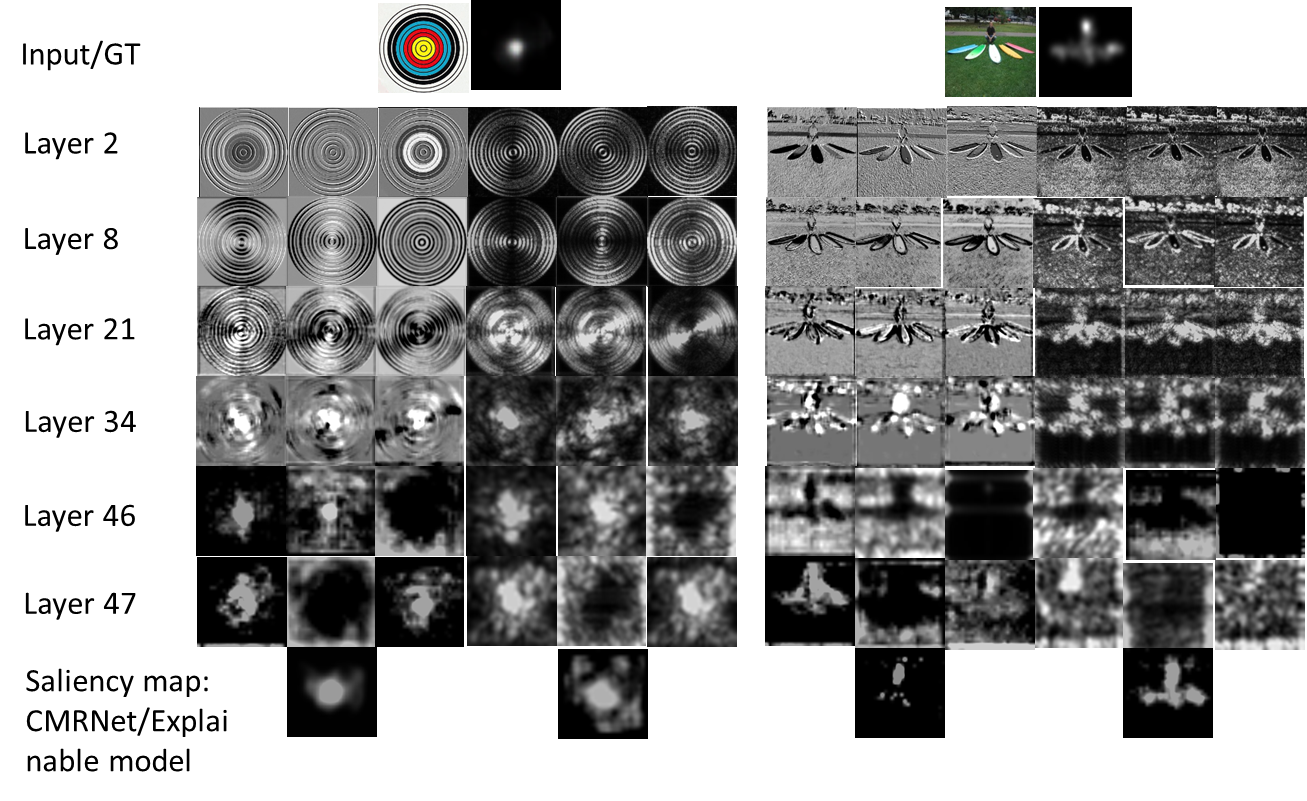}\label{fig:result_maps}}
\hfill
\subfloat[UNISAL]{\includegraphics[width=0.5\textwidth]{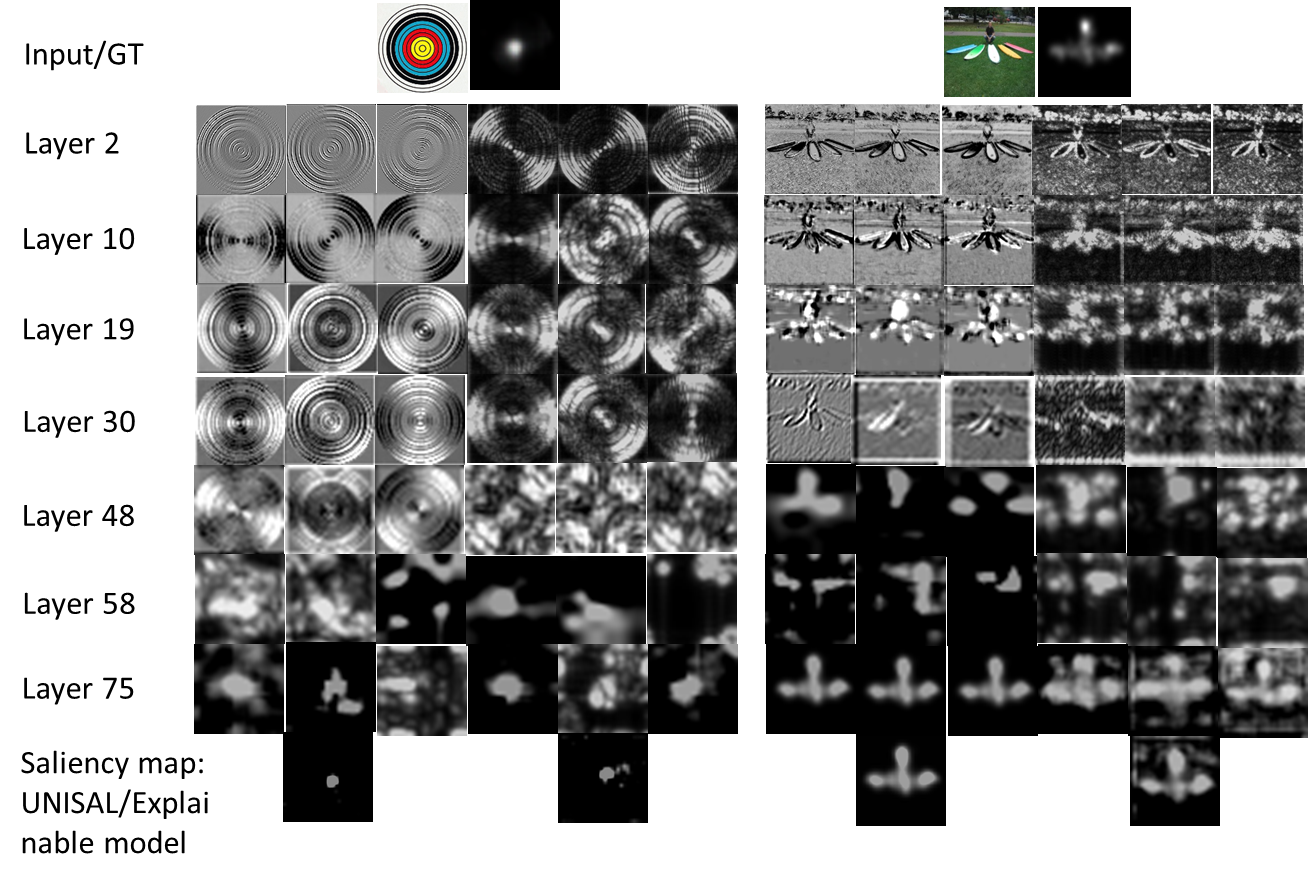}\label{fig:result_mapsdinet}}
\caption{Qualitative comparison of intermediate activation maps of our explainable saliency model with CMRNet and UNISAL on SALICON validation set.}
\label{fig:cmranddinet}
\end{figure*}

\subsection{Qualitative analysis}
Since UNISAL is found to be the best performing model on majority of the datasets, we show the quantitative results of UNISAL along with CMRNet. Fig. \ref{fig:cmranddinet} shows the activation maps generated by the explainable saliency models derived from both the CMRNet and UNISAL for the SALICON validation set. The first row contains both the input and the ground truth saliency map. From \textit{Layer 2} until \textit{Layer 47} in Fig. \ref{fig:result_maps} and from \textit{Layer 2} until \textit{Layer 59} in Fig. \ref{fig:result_mapsdinet} represent the activation maps. In Fig. \ref{fig:result_maps}, first and last three columns contain activation maps from CMRNet and the explainable model derived from CMRNet, respectively. In Fig. \ref{fig:result_mapsdinet}, it is the same except the first three columns are from UNISAL. The last row contains the final saliency map generated by the deep model (CMRNet or UNISAL) and their corresponding explainable saliency model.

In Fig. \ref{fig:result_maps}, our CMRNet has $50$ layers in it. Since it contains CMR blocks followed by a DIM and then the decoder, we provide the activation maps generated after each such prominent block. In Fig. \ref{fig:result_mapsdinet}, the UNISAL has $75$ deep layers in it. We provide the generated maps at intermediate layers of it. It can be seen that the reconstructed maps are considerably good in comparison. This is justified by evaluation metrics provided in Table \ref{table:salvalqualitative}. The explainable model has activation maps with extracted low level features in initial layers. While it became considerably denser in the latter layers wherein the thickness of edges gradually increases. If that thickness increases considerably to embrace the nearby thick edges, then the activation map will contain larger blobs. This leads to the extraction of dense features profoundingly indicating the model is learning the task of saliency prediction. Hence, in terms of the parameters of log-gabor filter bank, we found that their bandwidth gradually increases when we move deeper into the layers. 

Also, we see a good match of the final saliency maps generated by the deep models and their explainable model saliency maps with respect the ground truth. However, we see some false positives in saliency map of explainable model compared with deep model. As pointed earlier, it may happen because the former may not be able to imitate the nonlinear associations between the layers of the deep models. However, despite such intricacies in a deep model, our explainable model is able to mimic the deep models with a good amount of interpretation using the existing visual perception theories. In future, we would like to make it a completely independent and efficient module from DNN and outperform classical models. Thus, it reduces the usage of black-box DNNs and lessens the researchers’ efforts for debugging and fine tuning them.
\section{Conclusion}
In this work, we propose a technique to derive explainable deep visual saliency models from a deep neural network architecture based saliency models. We consider two state-of-the-art deep saliency models namely UNISAL and MSI-Net for our interpretation. We understand its intermediate activation maps and build a log-gabor filter bank ensuring that it encompasses all its parametric variations. For a given input image, we generate a set of log-gabor filter responses at the first layer. We consider block-based second order statistical descriptors to identify the activation maps in terms of the known filter responses. Then, we reconstruct them using a weighted linear combination of filter responses, thus eventually generating the final saliency map. Then, we build our own deep saliency model named cross-concatenated multi-scale residual block based network (CMRNet) with an efficient feature encoder. We perform the quantitative and qualitative evaluation of all these explainable models on three benchmark datasets. Its performance is found to be at par with the existing methods. Hence, we conclude that our explainable deep saliency model is generic and able to mimic any deep visual saliency model.

{\small
\bibliographystyle{ieee_fullname}
\bibliography{egbib}
}

\end{document}